\documentclass[final]{nesy2025} 

\usepackage{longtable}

\usepackage{booktabs}
\usepackage[load-configurations=version-1]{siunitx} 

\usepackage{framed}
\usepackage{subcaption}

\mathchardef\mhyphen="2D

\newcommand{\assum}{\rule[0.25cm]{0.5cm}{0.1pt}}

\theorembodyfont{\upshape}
\theoremheaderfont{\scshape}
\theorempostheader{:}
\theoremsep{\newline}

\title[Object-Centric Neuro-Argumentative Learning]{Object-Centric Neuro-Argumentative Learning}

\author{
\Name{Abdul Rahman Jacob}
\Email{abdul-rahman.jacob20@imperial.ac.uk}\\
\addr Imperial, London, UK
\AND
\Name{Avinash Kori}
\Email{a.kori21@ic.ac.uk}\\
\addr Imperial, London, UK
\AND
\Name{Emanuele {De Angelis}}
\Email{emanuele.deangelis@iasi.cnr.it}\\
\addr IASI-CNR, Rome, Italy
\AND
\Name{Ben Glocker}
\Email{b.glocker@imperial.ac.uk}\\
\addr Imperial, London, UK
\AND
\Name{Maurizio Proietti}
\Email{maurizio.proietti@iasi.cnr.it}\\
\addr IASI-CNR, Rome, Italy
\AND
\Name{Francesca Toni}
\Email{ft@ic.ac.uk}\\
\addr Imperial, London, UK
}
 
\begin{document}

\maketitle


\begin{abstract}
Over the last decade, as we rely more on deep learning technologies to make critical decisions, concerns regarding their safety, reliability and interpretability have emerged.
We introduce a novel Neural Argumentative Learning (NAL) architecture that integrates Assumption-Based Argumentation (ABA) with Object-Centric (OC) deep learning for image analysis. 
Our \emph{OC-NAL} architecture consists of neural and symbolic components. The former segments and encodes images into facts
, while the latter applies ABA learning to develop ABA frameworks enabling 
image classification. 
Experiments on synthetic data show that the OC-NAL architecture can be competitive with a state-of-the-art alternative. 
The code can be found at \url{https://github.com/AbdulRJacob/Neuro-AL}
\end{abstract}

\section{Introduction}
Over the last decade, AI, supported by deep learning, has become increasingly more  prevalent in our lives. However, 
as we rely more on deep learning technologies to make critical decisions, concerns regarding their safety, reliability and explainability naturally emerge
. Indeed, deep learning models, such as those used for image classification, are considered black boxes as their internal 
workings 
are not easily interpretable
, resulting in a possible lack of trust in their predictions. 

Motivated by the need for more interpretable 
image classifiers, 
we introduce a novel neuro-argumentative learning (NAL) architecture which generates symbolic representations in the form of assumption-based argumentation (ABA) frameworks \citep{RefWorks:RefID:17-phan2009assumption-based} from images, using objects identified in these images by Object-Centric (OC) methods~\citep{RefWorks:RefID:42-vita2020introduction}. The resulting ABA frameworks can be used to make predictions while allowing 
humans to follow a line of reasoning as to why the model made those predictions.

To generate ABA frameworks, our \emph{OC-NAL} architecture  uses ABA Learning~\citep{RefWorks:RefID:24-de2023aba,DPT-ECAI24}, a method that uses argumentation in a logic-based learning fashion to generate ABA frameworks which, with their accepted arguments, cover  given positive examples and do not cover given negative examples. 
OC-NAL also uses 
slot attention~\citep{RefWorks:RefID:3-francesco2020object-centric} as the underpinning OC method, to support
a granular understanding of input images in terms of the objects  they contain.
Overall, our OC-NAL architecture enables the extraction of meaningful properties and relationships between objects within the images, facilitating accurate classification with interpretable argumentation frameworks.

\begin{figure}[!t]
 \centering   \includegraphics[width=0.8\textwidth]{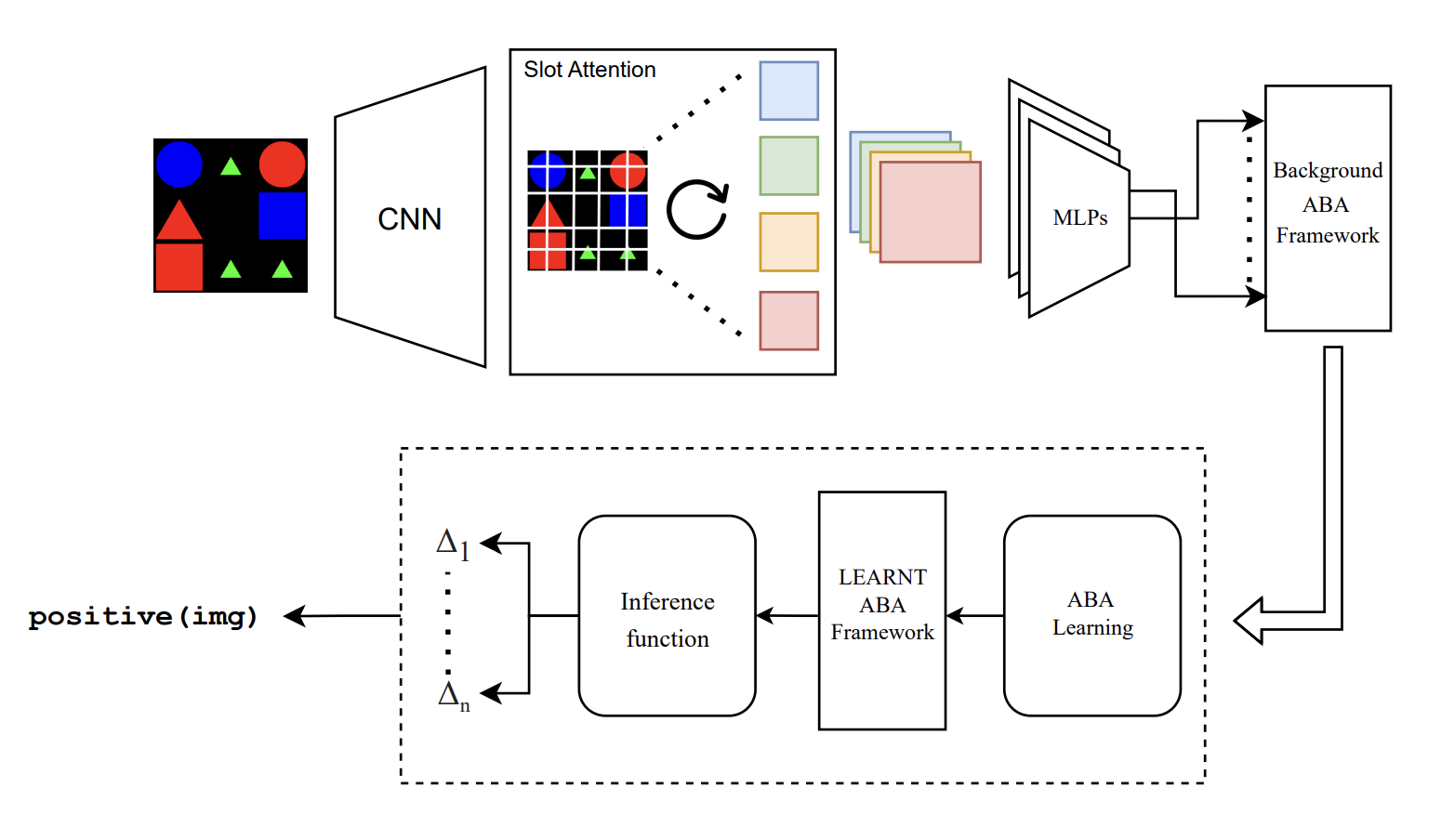}
    \caption{
    OC-NAL:  the input image (top left) is processed by slot attention to obtain objects (coloured squares) mapped by  
    MLPs into facts  fed into a Background ABA framework
    ; then ABA learning generates a Learnt ABA framework, which may admit several `extensions' {(i.e., sets of accepted arguments)} $\Delta_1, \ldots, \Delta_n$; inference therewith gives a classification (bottom left).}
    \label{fig:overview}
\end{figure}

\paragraph{Contributions} 
Overall, we make the following contributions: 1) we tailor slot-attention to generate factual background knowledge suitable to be injected in ABA Learning; 2) we combine slot-attention and ABA Learning into a pipeline architecture (that we term \emph{OC-NAL}) for neuro-argumentative learning; 3) we assess the performances of our method experimentally on (synthetic) image datasets, showing that it can be competitive against a baseline (Neuro-Symbolic Concept Learner (NS-CL)~\citep{RefWorks:RefID:34-mao2019neuro-symbolic}).

\section{Related Work}\label{sec:rel}
\cite{RefWorks:RefID:1-proietti2023roadmap} overview several approaches integrating logic-based learning with image classification. Specifically, 
NeSyFold~\citep{RefWorks:RefID:26-padalkar2023nesyfold:} uses the rule-based learning algorithm FOLD-SE-M to convert binarised kernels from a trained CNN to a set of ASP rules \citep{GeL91} with abstract predicates. It then uses semantic labelling to assign human-like concepts to predicates leading to global explanations 
of image classifications. Further, Embed2Sym~\citep{RefWorks:RefID:40-aspis2022embed2sym} 
uses clustered embeddings extracted by a neural network 
in combination with a symbolic reasoner encoded with predefined rules for explainable predictions. 
Some other  approaches integrate learning and argumentation for image classification, notably \cite{ECAI24imagesarg} classify images based on argumentative debates drawn from encoders
 and \cite{visualdebtes24}
 explain the outputs of 
 image classifiers with argumentative debates drawn from quantized features.
 \emph{No 
 existing approach 
uses argumentation with object-centric methods, as we do.} Indeed, to the best of our knowledge we are the first to 
propose such combination as a neuro-symbolic learning approach.

The closest approach to 
OC-NAL is  Neuro-Symbolic Concept Learner (NS-CL)~\citep{RefWorks:RefID:34-mao2019neuro-symbolic}, which we use as a baseline
. NS-CL uses object-centric learning via slot attention to identify objects within images, while its reasoning modules incorporate set transformers to extract and generate explanations
. These
are presented as grid-like representations, enabling users to understand the concepts applied in 
classification and derive further rules.
We use this approach as a baseline for the evaluation of our method, even though its outputs are at a different level of abstraction than our fully argumentative method.


\medskip

\section{Background}

In this section we briefly recall some notions that are the basis of our OC-NAL architecture.

\label{sec:pre}


\paragraph{Slot Attention}
Slot Attention~\citep{RefWorks:RefID:3-francesco2020object-centric} maps a set of $N$ input feature vectors $ z \in \mathbb{R}^{N \times d}$, of dimension $d$, obtained from an input image $x$, to a set of $K$
output vectors, of dimension $d_S (\leq d)$, $\hat{z} \in \mathbb{R}^{K \times d_S}$ that we refer to as slots.
The input features are projected with linear layers to create \textit{key} and \textit{value} vectors, represented by $\mathbf{k}$ and $\mathbf{v}$, respectively.
The slots are also projected with a linear layer, resulting in a \textit{query} vector $\mathbf{q}$. 
To simplify our exposition, later on, let $f_s$
denote the \textit{slot update} function, defined as:


\begin{equation}
    \hat{z}^{t+1} := f_s(z, \hat{z}^t) = \hat{A}\mathbf{v}, \, \hat{A}_{ij} := \frac{A_{ij}}{\sum_{l=1}^N A_{il}}, \,
    A := \text{softmax}\left(\frac{\mathbf{q} \mathbf{k}^T}{\sqrt{d_S}}\right)
    \label{eqn:slot_attention_update}
\end{equation}


\noindent
where $A \in \mathbb{R}^{K \times N}$ is the cross-attention matrix. 
The queries $\mathbf{q}$ in slot attention are a function of the slots $\hat{z}^t$, and are iteratively refined over $T$ iterations. The initial slots $\hat{z}^{t{=}0}$ are randomly sampled from a standard 
Gaussian~\citep{RefWorks:RefID:3-francesco2020object-centric}.

\paragraph{Assumption-Based Argumentation (ABA)} 

Assumption-Based Argumentation \citep{RefWorks:RefID:17-phan2009assumption-based} is a well-known symbolic formalism for modelling non-monotonic reasoning. 
An 
ABA framework~\citep{RefWorks:RefID:17-phan2009assumption-based}
is a tuple $ \langle \mathcal{L,R,A,} \assum \rangle$ where: 
(i) $\langle \mathcal{L,R} \rangle$ is a deductive system where $\mathcal{L}$ is the language and $\mathcal{R}$ is the set of (inference) rules; 
(ii) $\mathcal{A \subseteq  L}$ is the set of assumptions; 
(iii) $\assum$ is a total mapping from $\mathcal{A}$ to $\mathcal{L}$ where $\bar{a}$ is referred to a contrary of $a$ ($a \in \mathcal{A}$).
{In this paper, we consider \emph{flat} ABA frameworks, where assumptions are not heads of rules. 
Also, we assume that the elements of $\mathcal{L}$ are \emph{atoms} and,
for the sake of simplicity, we omit the language
as it can be derived from  $ \langle \mathcal{R,A,} \assum \rangle$.}
We illustrate with a simple example,
where, as in \citep{RefWorks:RefID:17-phan2009assumption-based}, we use schemata to write rules, assumptions and contraries, using the variable \texttt{A}.

\begin{example}\label{ex:ex1}
   A simple ABA framework for image classification is $ \langle \mathcal{R,A,} \assum \rangle$, where: 

   \vspace{-5mm}
   \[
    \begin{array}{ll}
              \mathcal{R} = \{ \quad  \rho_1: \texttt{circle(A) :- A=img\_1}, & \rho_2: \texttt{circle(A)\ :-\  A= img\_2},\;\; \\
     \phantom{\mathcal{R} = \{ \quad }\rho_3: \texttt{square(A)\ :-\ A= img\_2}, & \rho_4: \texttt{c\_1(A)\ :-\ circle(A),\ alpha(A)}, \\
     \phantom{\mathcal{R} = \{ \quad }\rho_5: \texttt{c\_alpha(A)\ :-\ square(A)} &
     \hspace{-1cm}\} \\[5pt]
     \mathcal{A} = \{ \texttt{alpha(A)} \} 
     \;\;\; \;\;\;\;\;
     \overline{\texttt{alpha(A)}} = \texttt{c\_alpha(A)}
    \end{array}
    \]
  Here, $\mathcal{R}$ is a set of \emph{rules}, each with a name $\rho_i$, a head following \texttt{:} and a body following \texttt{:-},
  $\mathcal{A}$ is a set of
  \emph{assumptions}, in this case consisting of a single assumption, with each assumption equipped with a \emph{contrary},
  in this case the contrary of $\texttt{alpha(X)}$ is $\overline{\texttt{alpha(X)}}$.
  {The intuirive meaning of the ABA framework is as follows: 
   images \texttt{img\_1} and \texttt{img\_2} contain a circle ($\rho_1$, $\rho_2$), 
   image \texttt{img\_2} contains a square ($\rho_3$), and 
   image \texttt{A} belongs to concept \texttt{c\_1}, if it contains a circle, unless it also contains a square ($\rho_4$, $\rho_5$).
  }
   \vspace{-2mm}
\end{example}
We define 
a \emph{fact} as a rule with distinct variables in the head and only equalities in the body.

To decide which conclusions may be drawn from an ABA framework, arguments and attacks between them are first obtained, then acceptance of arguments is determined using an extension-based semantics, in our case of \emph{stable extensions}~\citep{RefWorks:RefID:17-phan2009assumption-based}.
An \emph{argument} for the claim $c \in \mathcal{L} $ supported by $ A \subseteq \mathcal{A} $ and $ R \subseteq \mathcal{R}$ (denoted as $A \vdash_{R} c$) is a finite tree with nodes labelled by sentences in $\mathcal{L}$ or by $\tau$ denoting \textit{true}, the roots labelled by $c$, the leaves either \textit{true} or assumptions in $A$, and non-leaves $c'$ with, as children, the elements of the body of some rule in $R$ with the head $c'$.
An argument $A \vdash_{R} c$
\emph{attacks} an argument $A' \vdash_{R'} c'$ iff there is an assumption $a \in A'$ such that $\overline{a}=c$.
A set of arguments $E$ is \emph{stable} iff
the set is conflict-free (i.e. no argument in $E$ attacks an an argument also in $E$) and for every argument not in $E$ there is an argument in $E$ attacking it.
We illustrate these notions with the earlier example.

\begin{example}\label{ex:ex2}
The following arguments  can be obtained (amongst others) from the ABA framework in example \ref{ex:ex1}:


\vspace{-7mm}
\[
\begin{array}{ll}

\{\texttt{alpha(img\_1)} \} \ \vdash_{\{\rho_1,\rho_4\}} \texttt{c\_1(img\_1)}, 
&
\{\texttt{alpha(img\_2)} \} \ \vdash_{\{\rho_2,\rho_4\}} \texttt{c\_1(img\_2)},\\

\{ \} \ \vdash_{\{\rho_2\}} \texttt{circle(img\_2)}, 
&
\{\} \ \vdash_{\{\rho_3,\rho_5\}} \texttt{c\_alpha(img\_2)}.


\end{array}
\]


\noindent
Intuitively, each argument is a deduction from (possibly empty) sets of assumptions (the premises) to claims (e.g. $\texttt{c\_1(img\_1)}$ for the first argument), using sets of rules. 
Attacks between arguments result from undercutting assumptions in the premises of arguments.
Here, 
the fourth argument attacks the second, as the former is a deduction of the contrary of the assumption occurring in the premise of the latter. 
The third and fourth arguments belong to the single stable extension admitted by this simple ABA framework, 
as they cannot be attacked by any other arguments. 
\end{example}



\paragraph{Learning ABA Frameworks} 
We use the ASP-ABALearn method by \cite{RefWorks:RefID:24-de2023aba,DPT-ECAI24}.
This takes in input
a Background ABA framework (admitting at least one stable extension), sets $\mathcal{E}^+$ and $\mathcal{E}^-$ of positive and negative examples {(i.e., atoms obtained from labelled images), } respectively, and returns in output a Learnt ABA framework (admitting at least one stable extension) such that all positive examples are accepted in all stable extensions and no negative example is accepted in all the stable extensions.
Computationally, ASP-ABALearn
leverages the fact that flat ABA frameworks (where assumptions 
cannot be claims of arguments supported by other assumptions) can be mapped to logic programs. This is done by replacing each assumption $\alpha(X)$ with $not \ p(X)$ where  $\overline{\alpha(X)} = p(X)$
.


\section{OC-NAL Architecture}\label{sec:main}
We will now explain the OC-NAL architecture shown in Figure \ref{fig:overview},
by detailing the neural and symbolic components and their training, as well as inference post-learning.


\paragraph{Inputs}
The OC-NAL architecture accepts a dataset $D \subseteq X\times Y \times L$ of 
labelled images, where $X$ is the given set of images, $L=\{c_1, c_2\}$ is a set of classes, and 
$Y=\{0,1\}^{K \times (P + 1 )}$, for
$K$ 
the total number of objects/slots that may occur in images, and $P$ the total number of \emph{properties}
that each of these objects may have (we consider an extra property for characterising the absence of objects). 
As a simple example, for images such as the one in Figure~\ref{fig:overview},  $K=10$ (as there are a maximum of 9 objects in each such image plus the background) and $P=8$ (3 for the shapes, 3 for the colours, 2 for sizes).
We assume that $D$ is not noisy
.
$L$ is a set of two alternative classes.
$Y$ is a metadata consisting of one-hot 
encodings, each representing all objects and their corresponding properties in an image.
Note that the neural component of the architecture disregards the labels in $L$,
focusing instead on the images in $X$ in a  weakly supervised manner, while the symbolic component disregards the image itself, using instead its abstraction drawn from the neural component.




\paragraph{Neural Component}
This uses a Convolutional Neural Network (CNN), slot attention and a set of multi-layer perceptrons (MLP) to convert a given input image $x$ into facts for the symbolic component, amounting specifically to the Background Knowledge for ABA Learning or, during inference, facts to be added to the generated ABA framework.

First, the input image is converted into features $z$ using a CNN, which is further used by the slot attention model trained using the process described in Section~\ref{sec:pre} to produce slots~$\hat{z}$.
%

Then, each slot $\hat{z}$ is passed through the MLPs to extract the properties of each object. 
To identify both continuous and categorical properties, we use MLPs of two types: classification MLPs, which use a softmax activation in the final layer to predict the most likely attribute for a given slot and regression MLPs, to predict the location of objects and determine whether each given slot has attended to a real object in the image. 
The results of these predictions are then concatenated to form the final prediction $\hat{y}$ for the input image, with the corresponding ground truth $y\in Y$. 
This component is trained with weak-supervision and is optimised by minimising the loss function:  
\vspace{-3mm}

\begin{equation}
\text{MSE}(x, \hat{x}) + \alpha \min_{\tau \in S_K} \sum_{j=1}^{P} \text{BCE}(y_j, \tau(\hat{y})_j) 
\label{eqn:loss}
\end{equation} 

\vspace{-2mm}

\noindent which encapsulates both training objectives. The first is the mean square error (MSE) between the input image $x$ and the (reconstructed) image $\hat{x}$, ensuring the
reconstruction quality (from the slots) of the model. The second is the binary cross entropy (BCE) between the ground truth label $y$ and the predicted label $\hat{y}$,
where $y_j$ corresponds to a particular property in vector $y$ and $\tau(\hat{y})$ is 
the prediction for a permutation of the $K$ objects, drawn from
$S_K$, which denotes the set of all such permutations of $K$ objects. 
Intuitively, we compare
each of the permutations with the ground-truth representation (with a specific order).
Given the equivariance property of slot attention, we need to first align the slots before applying the BCE loss. To circumvent this,  we used the Hungarian matching algorithm~\citep{RefWorks:RefID:43-kuhn1955hungarian}. 
Finally we balance both loss terms with the hyperparameter $\alpha$.


\paragraph{Symbolic Component}
This receives the output predictions from the neural component and transforms them into facts for the Background ABA Framework taken in input by the ASP-ABALearn algorithm.
It also uses the input labels in $L$ to obtain appropriate positive and negative examples in $(\mathcal{E}^+, \mathcal{E}^-)$. 
This is accomplished by aggregating the slots and performing K-means clustering. The number of clusters corresponds to the desired number of examples, and an image is chosen from each cluster as a representative positive or negative example. We also check the confidence of each prediction and prune off any image below a certain threshold. 

The slot predictions are then passed to concept embedding functions which use a dictionary (for the $K$ objects and the $P$ properties) to convert the raw predictions to ABA facts. For each image and object, an identifier is given in the form of an atom $\texttt{image(img\_i)}$ and a constant $\texttt{object\_i}$ respectively. Then, for each slot prediction, we take the argmax to identify the properties in the dictionary which are attributed to the object. We encode this as a fact, e.g. $\texttt{blue(object\_i)}$. 

Once all images and objects are encoded into the Background ABA Framework, we generate the ASP-ABALearn command $\texttt{aba\_asp(`filename.aba',}$ $\texttt{e\_pos,e\_neg).}$
%
This specifies which images are positive/negative, using $\texttt{e\_pos}$/$\texttt{e\_neg}$ as the encodings of $\mathcal{E}^+/\mathcal{E}^-$, as discussed earlier. The ABA-ASPLearn algorithm is then run to produce a Learnt ABA Framework. 


\paragraph{Inference}
At inference time, we run a slightly different pipeline to obtain a final classification for each 
unseen input image. 
Specifically: 
\begin{enumerate}
    \item 
We pass the image through the neural component to obtain predictions of the objects and their properties therein. These are subsequently converted into facts as during training.
\item 
We then create an ABA Framework which contains these facts, the rules learnt via ASP-ABALearn 
and any extra background knowledge.
\item The stable extensions of this ABA framework are then computed (using a straightforward mapping into ASP, and then using an ASP solver. specifically we use Clingo~\citep{RefWorks:RefID:44-gebser2014clingo}). 
\item Depending on the ABA framework, we may obtain more than one stable extension. 
The prediction boils down to checking whether the atom sanctioning that the input image belongs to concept $c_1$ 
is a member of all the stable extensions (i.e., it is a \emph{cautious} consequence of the Learnt ABA framework).
\end{enumerate}


\section{Experimental Evaluation}\label{sec:eval}
We conducted experiments on the OC-NAL architecture 
to answer the following questions:

\begin{itemize}
    \item[\textbf{Q1}:]  How well can the 
Neural Component identify/predict object properties?

\item[\textbf{Q2}:] How well can the OC-NAL architecture learn ABA frameworks which describe the latent rules in images 
so that they be used for classification
?

\item[\textbf{Q3}:]  How well does our OC-NAL architecture scale w.r.t. the number of examples used in the Symbolic Component and the complexity of the latent rules?

\end{itemize}


\paragraph{Experiments} To address these questions, we defined various binary classification tasks using our 
adaptation of the SHAPES dataset~\citep{shapesdataset23}
. This dataset was generated by a tool that processed ASP rules to create images conforming to them. We defined 6 rules, as detailed in Figure \ref{table:shape_dataset_rules}, to form the dataset. Each binary classification task aimed to distinguish between positive and negative instances for each class. This dataset served as a baseline to evaluate the viability of using argumentation to reason in an object-centric way.
We also defined a multi-class classification task on the CLEVR dataset using CLEVR-Hans3~\citep{RefWorks:RefID:2-stammer2021concept:} which splits CLEVR images into 3 classes based on the following concepts c1: \texttt{Large (Gray) Cube and Large Cylinder}, c2: \texttt{Small metal Cube and Small (metal) Sphere}, c3: \texttt{Large blue Sphere and Small yellow Sphere}. The goal of this task was for OC-NAL to generate Learnt ABA frameworks that could differentiate these classes. 

\begin{figure}[t]
\centering
\resizebox{\columnwidth}{!}{
\begin{tabular}{@{\hspace{3pt}}r@{\hspace{3pt}}l@{\hspace{3pt}}l}
\toprule
\multicolumn{3}{c}{SHAPES Dataset ASP Rules} \\ \midrule
\texttt{s1(A)} & :- & \texttt{image(A), in(A,B), square(B), blue(B).} \\
\texttt{s2(A)} & :- & \texttt{image(A), in(A,B), triangle(B), small(B), green(B).} \\
\texttt{s3(A)} & :- & \texttt{image(A), in(A,B), in(A,C), triangle(B), blue(B), circle(C), red(C), large(C).} \\
\texttt{s4(A)} & :- & \texttt{image(A), in(A,B), in(A,C), circle(B), red(B), square(C), blue(C), above(B,C).} \\
\texttt{s5(A)} & :- & \texttt{image(A), in(A,B), in(A,C), triangle(B), red(B), circle(C), green(C), left(B,C).} \\
\texttt{s6(A)} & :- & \texttt{not exception(A), image(A).}\\
\texttt{exception(A)} & :- & \texttt{image(A), in(A,B), circle(B), blue(B).}\\ \bottomrule
\end{tabular}
}

\caption{ASP rules used to generate the 6 classes for the SHAPES dataset. We generated 3K images for each rule half of which was the negative instance of the rules. We then took 500 positive and negative splits for each rule as testing data. In total, we had 12K images for training and 6K for testing.
}
\label{table:shape_dataset_rules}
\end{figure}


\paragraph{Setup} We trained the OC-NAL architecture in two stages as described in Section~\ref{sec:main}. The Neural Component was trained using the full datasets for 1000 epochs with hyperparameter $\alpha = 0.35$. The Symbolic Component used 10 positive and 10 negative examples from the datasets to obtain the Learnt ABA framework. Regarding the CLEVR-Hans3 task (which was multi-classed), we executed the Symbolic Component twice, the first run to distinguish
c3 images from c1 and c2 and the second run to distinguish between c1 and c2, thus producing two frameworks. We compare the results with a ResNet~\citep{RefWorks:RefID:45-he2016deep} and the NS-CL.

\begin{figure*}
    \centering
    \includegraphics[width=\textwidth]{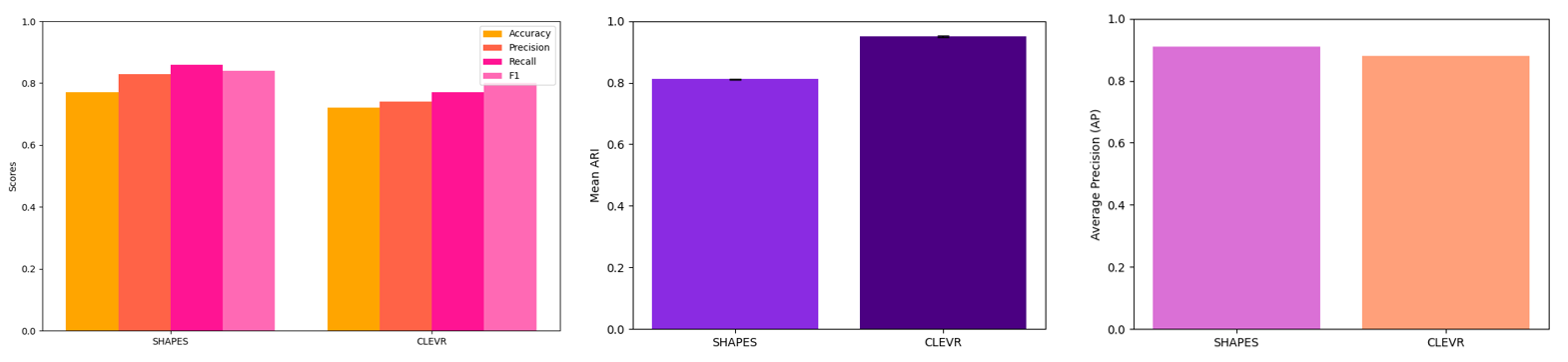}
    \caption{\textbf{(Left)} Standard machine learning evaluation metrics of attribute classification on datasets SHAPES and CLEVR \textbf{(Center)} Mean Adjusted Rand Index on datasets \textbf{(Right)} Average Precision on datasets.
    }
    \label{fig:neural_eval}
\end{figure*}

\paragraph{[Q1] Object Prediction} Figure~\ref{fig:neural_eval} shows that the Neural Component effectively segmented images into their constituent objects and accurately predicted each object's property. We evaluated its localisation and segmentation capabilities using the Adjusted Rand Index, which measures similarity between clusters (with clusters representing objects and data points representing pixels), and the Average Precision metric. The scores ranged from 0.80 to 0.95, indicating consistent performance regardless of the number of objects present in each dataset.

We also observed high scores in standard 
metrics, indicating that the MLPs accurately predicted each object's properties. The F1 score exceeded 70\%, suggesting that the model effectively minimises false positives and negatives, ensuring the symbolic component contains facts that accurately represent the image. Results on the CLEVR dataset were slightly worse than those on the SHAPES dataset, likely due to CLEVR objects having a larger number of properties. This suggests that the number of properties can affect prediction performance.


\paragraph{[Q2] Classification
} From Table \ref{table:shapes_ml_metrics} we observe that the Learnt ABA frameworks performed well on the binary classification tasks for SHAPES, achieving near-perfect scores on most tasks. Among the metrics, we saw that recall was consistently the lowest. This is likely due to errors propagated from the Neural Component, which resulted in facts that did not accurately represent the images, causing some instances to be misclassified.

\begin{figure}[t]
    \centering
    \begin{minipage}{0.5\textwidth}
        \centering
        \resizebox{\textwidth}{!}{
            \begin{tabular}{l c c c c}
                \toprule
                & Accuracy & Precision & Recall & F1-Score \\
                \midrule
                OC-NAL & $69.1 \pm 0.3$ & $70.0 \pm 0.1$ & $69.1 \pm 0.2$ & $68.0 \pm 0.2$ \\
                ResNet & $65.2 \pm 0.3$ & $66.2 \pm 0.4$ & $65.2 \pm 0.2$ & $61.0 \pm 0.2$ \\
                NS-CL & $84.7 \pm 0.1$ & $86.1 \pm 0.2$ & $84.7 \pm 0.2$ & $84.0 \pm 0.2$ \\
                \bottomrule
            \end{tabular}
        }
        \caption*{Table 2: Standard evaluation metrics for the performance of OC-NAL, ResNet, and NS-CL on the CLEVR-Hans3 task.}
        \label{table:clevr-hans3}
    \end{minipage}%
    \hfill
    \begin{minipage}{0.48\textwidth}
        \centering
        \includegraphics[width=0.9\textwidth,trim={0 0.4cm 0 0.2cm},clip]{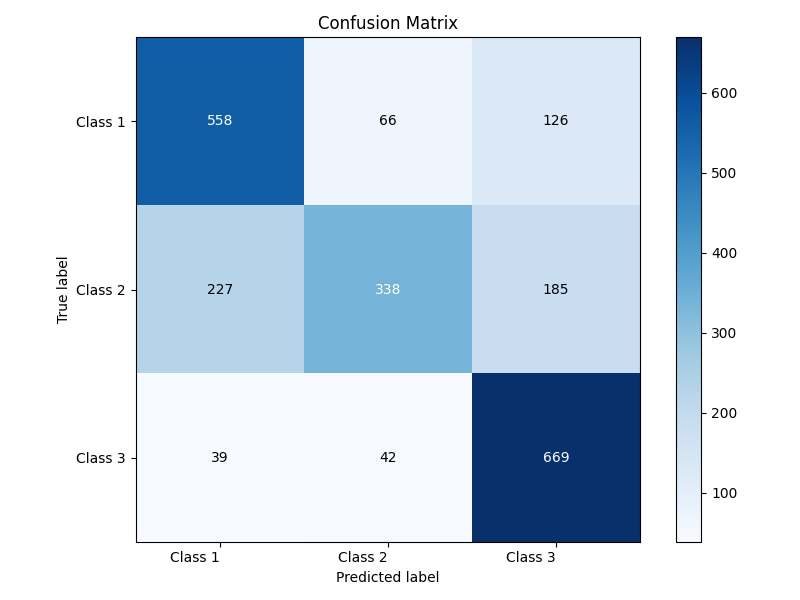}
        \caption{Confusion Matrix of OC-NAL on the CLEVR-Hans3 task.}
        \label{fig:cm}
    \end{minipage}
\end{figure}

\noindent However, we saw the opposite 
when predicting classes s4 and s5,  with significantly lower precision. We believe that this is due to the fact that in those tasks ASP-ABALearn learnt rules capturing (some of) the exceptions rather than the full concepts. For example, s1 images were defined in the dataset (see Figure~\ref{table:shape_dataset_rules}) as those with blue squares. However, the learnt framework
defined $\texttt{s\_1}$ as images with squares that are not red and not green (see \texttt{c\_alpha2} in
Figure \ref{fig:aba_outputs}). Despite being semantically equivalent, this reasoning made interpretation difficult and impacted the generation of more complex rules. Consequently, the F1 scores for these tasks dropped significantly due to the inability to capture all rule exceptions, leading to many false positives and thus lower precision.
\\
We encountered a similar outcome with the learnt ABA frameworks generated for the CLEVR-Han3 tasks. These only partially captured the rules defining each class. For instance, the framework for c1 stated that these images contain cubes that are not small. However, this rule failed to distinguish some c2 images, which could also contain cubes that are not small. As a result, many c2 images were incorrectly classified as c1, as shown in the confusion matrix (see Figure~\ref{fig:cm}).

Despite this issue, the learnt ABA frameworks were able to capture most of the facts for classifying c3 images, leading to an F1 score of 0.68 (see Table~2). This outperformed the ResNet baseline
, though it was 
worse than NS-CL, which scored 
above 0.80\%. The superior performance of NS-CL could be attributed to its use of a set transformer for classification, rather than relying solely on symbolic reasoning.

\begin{figure}
    \begin{framed}
    \begin{minipage}[t]{\textwidth}
    \texttt{\% Learnt Rules} \\[1pt] 
    \texttt{%
        s\_1(A) :- in(A,B), square(B), alpha\_2(B,A).\\
        c\_alpha\_2(A,B) :- image(B), red(A).\\
        c\_alpha\_2(A,B) :- image(B), green(A). 
        }   
    \end{minipage}  
    \end{framed}
 
\caption{
Rules in the ABA framework generated by our OC-NAL architecture 
for SHAPES (class s1). 
Here, {\tt alpha\_2} is an assumption,
with $\overline{{\tt alpha\_2(A,B)}}$={\tt c\_alpha\_2(A,B)}.} 
\label{fig:aba_outputs}
\end{figure}

\begin{figure}
    \begin{framed}
    \begin{minipage}[t]{\textwidth}
    \texttt{\% Learnt Rules} \\
    \texttt{%
        c\_1(A) :- in(A,B), cube(B), alpha\_2(B,A).\\
        c\_alpha\_2(A,B) :- small(A), image(B).
        }   
    \end{minipage}
    \end{framed}
 
\caption{
Rules in the ABA framework generated by our OC-NAL architecture 
for SHAPES (class s5).
Here, 
{\tt alpha\_2} is an assumption,
with $\overline{{\tt alpha\_2(A,B)}}$={\tt c\_alpha\_2(A,B)}.} 
\end{figure}

\begin{figure}        

    \begin{framed}
    \begin{minipage}[t]{\textwidth}
    \texttt{\% Learnt Rules} \\
    \texttt{%
        c\_3(A) :- in(A,B), sphere(B), alpha\_2(B,A).\\
        c\_alpha\_2(A,B) :- brown(A), image(B).\\
        c\_alpha\_2(A,B) :- green(A), image(B).\\
        c\_alpha\_2(A,B) :- cyan(A), image(B).\\
        c\_alpha\_2(A,B) :- red(A), image(B).\\
        c\_alpha\_2(A,B) :- large(A), image(B).\\
        c\_alpha\_2(A,B) :- blue(A), image(B).\\
        c\_alpha\_2(A,B) :- gray(A), image(B).
}
        \end{minipage}      
    \end{framed}
    \caption{Rules in the ABA framework generated by our OC-NAL architecture 
    for CLEVR, differentiating class c3 from c1 and c2.
    Here, 
    {\tt alpha\_2} is an assumption,
with $\overline{{\tt alpha\_2(A,B)}}$={\tt c\_alpha\_2(A,B)}.
} 

\end{figure}

\begin{table}
    \centering 
        \begin{tabular}{@{\hspace{3pt}}c@{\hspace{6pt}}c@{\hspace{6pt}}c@{\hspace{6pt}}c@{\hspace{6pt}}c@{\hspace{6pt}}c@{\hspace{3pt}}}
        \toprule
        Predicted class & Accuracy & Precision & Recall & F1-Score \\
        \midrule
        s1 & $99.0 \pm 0.0$ & $100.0 \pm 0.0$ & $97.5 \pm 0.0$ & $99.0 \pm 0.2$ \\
        s2 & $96.0 \pm 0.0$ & $99.0 \pm 0.0$ & $93.5 \pm 0.0$ & $96.0 \pm 0.2$ \\
         s3 & $98.0 \pm 0.0$ & $100.0 \pm 0.0$ & $97.5 \pm 0.2$ & $98.0 \pm 0.0$ \\
         s4 & $75.0 \pm 3.0$ & $61.0 \pm 5.2$ & $98.5 \pm 0.4$ & $75.0 \pm 0.2$ \\
         s5 & $86.0 \pm 4.1$ & $77.0 \pm 3.8$ & $96.5 \pm 0.2$ & $84.0 \pm 0.2$ \\
         s6 & $99.0 \pm 0.0$ & $98.0 \pm 0.0$ & $100.0 \pm 0.0$ & $99.0 \pm 0.0$\\
        \bottomrule
        \end{tabular}

    \bigskip
    \caption{Standard evaluation metrics denoting how well OC-NAL can distinguish between positive and negative instances of rules present in SHAPES images}
    \label{table:shapes_ml_metrics}
\end{table}



\paragraph{[Q3] Scalability} During our experiment, we found that the symbolic component, specifically ASP-ABALearn, faced some scalability issues as the execution time grew significantly as we increased the number of examples. This could be due to a larger search space generated when looking for ABA frameworks whose extensions cover all positive examples and none of the negative examples. It may happen that this search is unsuccessful and ASP-ABALearn halts with failure.

Our experiments also showed that the non-determinism present when generalising rules led to variability in both the quality of the results and the system's execution time. This variability was amplified by the size of the Background ABA framework. Additionally, we observed that longer execution task times negatively impacted the quality of the results, as heavily nested rules caused the frameworks to overfit to the example set.

\section{Conclusions and Future Work}\label{sec:concl}

We have proposed a novel neuro-argumentative learning (OC-NAL) method for image classification, in the spirit of \citep{RefWorks:RefID:1-proietti2023roadmap}, integrating slot attention for object identification in images by \cite{RefWorks:RefID:42-vita2020introduction}, and the ASP-ABALearn  implementation of ABA Learning by \cite{RefWorks:RefID:24-de2023aba,DPT-ECAI24}. The proposed framework follows a faithful and human-understandable reasoning process. We empirically demonstrate 
that our approach can be effective 
via experiments with the resulting architecture on datasets of synthetic images.

Overall, we believe that this work has contributed some insights into the potential of 
argumentation in the space of image classification, demonstrating that an object-centric approach combined with ABA is a viable 
approach for neuro-symbolic learning. 
At the same time, several avenues for future work remain open.
Specifically, 
we plan to extend our framework to deal with real images, rather than synthetic images.
Also,  we plan to explore novel instances of our general architecture by considering other forms of slot-attention, e.g. the method of~\citep{KoriICLR24}
, and bespoke forms of ABA Learning suited to our setting.
Furthermore, it would be interesting to explore variants of our approach where slot-attention and ABA learning are trained together, in an end-to-end fashion.  
Finally, we plan to explore the explainability of classification with our form of NAL, especially in comparison with the  NS-CL baseline~\citep{RefWorks:RefID:34-mao2019neuro-symbolic}).

\newpage 
\section*{Acknowledgements}



We thank support from the Royal Society, UK
(IEC\textbackslash R2\textbackslash 222045 - International Exchanges 2022).
Toni was partially funded by the ERC under
the EU’s Horizon 2020 research and innovation 
programme (grant agreement No. 101020934) and 
by J.P. Morgan and the Royal Academy of Engineering,
UK, under the Research Chairs and Senior Research Fellowships scheme.
Kori was supported by UKRI through the CDT in Safe and Trusted Artificial Intelligence.
This paper has also been partially supported by the 
Italian MUR PRIN 2022 Project DOMAIN 
(2022TSYYKJ, CUP B53D23013220006, PNRR M4.C2.1.1) funded by the European Union – NextGenerationEU,
and by the PNRR MUR project PE0000013-FAIR - Future Artificial Intelligence Research (CUP B53C22003630006).
De Angelis and Proietti are members of the INdAM-GNCS research group.



%

\bibliography{main}

\end{document}